\documentclass[letterpaper, 10 pt, conference]{ieeeconf}

\IEEEoverridecommandlockouts

\usepackage{etextools}
\usepackage{amssymb}
\usepackage{float}
\usepackage{tabto}
\usepackage{makeidx}
\usepackage{amsmath}
\usepackage{bbm}
\usepackage{epsfig}
\usepackage{epsf}
\usepackage{psfrag}
\usepackage{verbatim}
\usepackage{color}
\usepackage{multirow}
\usepackage{tabularx}
\usepackage{booktabs}
\usepackage[tight,footnotesize]{subfigure}
\usepackage{array}
\usepackage{soul}
\usepackage{footnote}
\usepackage{cite}
\usepackage{dblfloatfix}
\usepackage{changepage}

\usepackage{color, colortbl}
\usepackage[colorlinks,bookmarksnumbered,citecolor=orange,urlcolor=orange]{hyperref}
\usepackage{graphicx}
\graphicspath{{Figures}}
\DeclareGraphicsExtensions{.pdf,.png}
\usepackage{bigstrut}
\usepackage[english]{babel}

\usepackage{csquotes}

\usepackage[T1]{fontenc}
\usepackage{algorithm, setspace}
\usepackage{algpseudocode}
\usepackage{url}
\usepackage{multirow}
\usepackage{float}
\usepackage{xcolor}
\usepackage{hyperref}
 \hypersetup{
     colorlinks=true,
     linkcolor=orange,
     filecolor=orange,
     citecolor=orange,      
     urlcolor=orange,
     }

\newcommand{\p}[1]{\smallskip \noindent \textbf{{#1}.}}
\newcommand{\eq}[1]{Equation~(\ref{eq:#1})}
\newcommand{\fig}[1]{Figure~\ref{fig:#1}}
\usepackage{balance}

\title{\LARGE

Waypoint-Based Reinforcement Learning for Robot Manipulation Tasks}

\author{Shaunak A. Mehta, Soheil Habibian, and Dylan P. Losey
\thanks{This work is supported in part by NSF Grants \#2129201 and \#2205241. \newline The authors are with the Collaborative Robotics Lab (\href{https://collab.me.vt.edu/}{Collab}), Dept. of Mechanical Engineering, Virginia Tech, Blacksburg, VA 24061.
\newline Corresponding author's email: \texttt{mehtashaunak@vt.edu}}
}

\begin{document}
\maketitle

\begin{abstract}

Robot arms should be able to learn new tasks.
One framework here is reinforcement learning, where the robot is given a reward function that encodes the task, and the robot autonomously learns actions to maximize its reward.
Existing approaches to reinforcement learning often frame this problem as a Markov decision process, and learn a policy (or a hierarchy of policies) to complete the task.
These policies reason over hundreds of fine-grained actions that the robot arm needs to take: e.g., moving slightly to the right or rotating the end-effector a few degrees.
But the manipulation tasks that we want robots to perform can often be broken down into a small number of high-level motions: e.g., reaching an object or turning a handle.
In this paper we therefore propose a \textit{waypoint-based} approach for model-free reinforcement learning.
Instead of learning a low-level policy, the robot now learns a trajectory of waypoints, and then interpolates between those waypoints using existing controllers.
Our key novelty is framing this waypoint-based setting as a sequence of multi-armed bandits: each bandit problem corresponds to one waypoint along the robot's motion.
We theoretically show that an ideal solution to this reformulation has lower regret bounds than standard frameworks.
We also introduce an approximate posterior sampling solution that builds the robot's motion one waypoint at a time.
Results across benchmark simulations and two real-world experiments suggest that this proposed approach learns new tasks more quickly than state-of-the-art baselines. 
See videos here: \url{https://youtu.be/MMEd-lYfq4Y}

\end{abstract}


\section{Introduction} \label{sec:intro}

\begin{figure}[t]
	\begin{center}
 		\includegraphics[width=1\columnwidth]{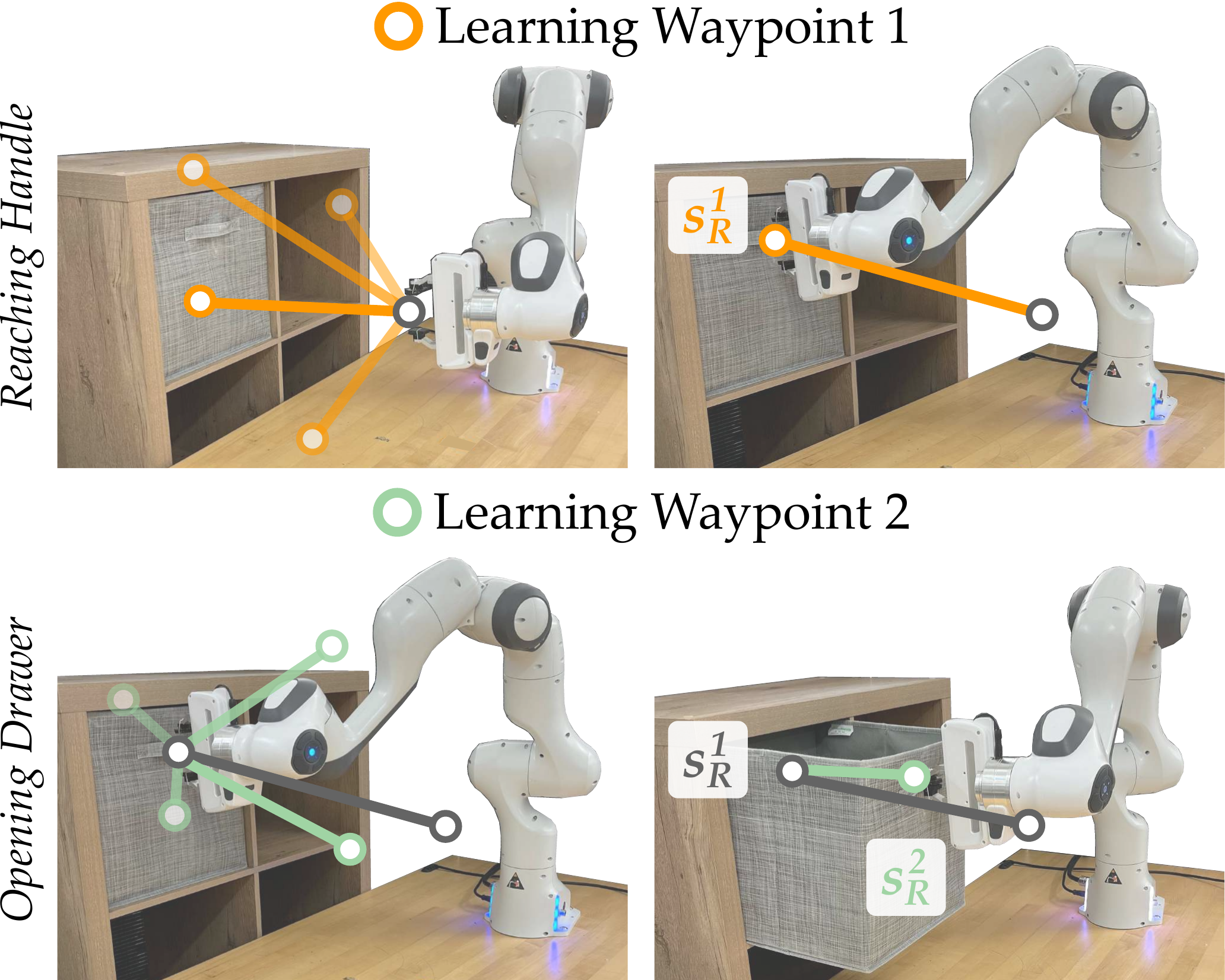}
		\caption{Our waypoint-based approach for model-free reinforcement learning in manipulation tasks. The robot arm learns where to place the next waypoint to maximize its reward by solving a multi-armed bandit. We then freeze the learned models for waypoint $i$, and repeat the process for waypoint $i+1$. This approach learns the desired task across a distribution of initial states; i.e., the location and angle of the drawers can change at each interaction.}
		\label{fig:front}
	\end{center}
	\vspace{-1em}
\end{figure}

Robots often need to learn behaviors that optimize a reward function.
For instance, in \fig{front} the robot's objective is to open a drawer.
To learn how to open this drawer the robot rolls-out behaviors in the environment and determines which actions increase its reward (i.e., which actions open the drawer).
Existing approaches often solve this reinforcement learning problem by constructing a policy.
Policies map states to \textit{fine-grained actions}: e.g., when the arm is near the handle it moves slightly forward, and when the arm is holding the handle it pulls slightly backwards.
In practice, this means the robot must reason about hundreds of low-level decisions throughout the task (i.e., each action of the policy).
But at a high-level the drawer task can be broken down into three stages: reaching the handle, grasping the handle, and sliding the drawer open.
Hence, the robot could solve this task --- and maximize its given reward --- by learning these \textit{high-level waypoints} and interpolating between them.

In this paper we propose a waypoint-based approached for model-free reinforcement learning.
Our work focuses on robot arms: inspired by recent research \cite{belkhale2023hydra, akgun2012keyframe, shi2023waypoint}, we recognize that many robot manipulation tasks can be broken down into a sequence of waypoints.
Moving between these waypoints is well understood: robot arms can leverage low-level controllers to track a reference trajectory \cite{spong2020robot, haddadin2016physical}.
But to learn the correct waypoints in the first place, the robot must trade-off between exploring the workspace and exploiting high performing areas.
Instead of formalizing this reinforcement learning problem as a Markov decision process --- a standard approach in robotics \cite{han2023survey} --- our insight is that:
\begin{center}\vspace{-0.3em}
\textit{Each waypoint is a continuous multi-armed bandit problem, where the \emph{arm} is the waypoint the robot selects and the \emph{reward} for moving to that waypoint is unknown a priori}.
\vspace{-0.3em}
\end{center}
Using this insight we introduce a method that builds trajectories \textit{one waypoint at a time} to maximize the robot's reward.
When applied to \fig{front}, our algorithm causes the robot to iteratively sample a waypoint and then roll-out a trajectory that terminates at that waypoint (e.g., trying a point to the left of the drawer).
Based on the measured rewards for this rolled-out trajectory, the robot updates its estimate of the waypoint reward and optimizes the current waypoint (e.g., correctly reaching the drawer handle during the next roll-out).
The robot then saves what it has learned about waypoint $i$ and repeats this process for waypoint $i+1$.
As we will show in our experiments, this leads to robots that learn to open the drawer in fewer interactions than state-of-the-art baselines.

Overall, we make the following contributions:

\p{Formulating as Sequential Multi-Armed Bandits}
We use waypoints to write model-free reinforcement learning for robot manipulation tasks as a sequence of continuous multi-armed bandits, where each bandit problem corresponds to one waypoint along the robot's learned trajectory.
We theoretically demonstrate that this proposed formulation can have lower regret bounds than existing frameworks.

\p{Learning Waypoints via Posterior Sampling}
We next introduce an algorithm that approximately solves the sequence of multi-armed bandits through posterior sampling.
Our method maintains an ensemble of models to estimate the reward function for a given waypoint; at each interaction the robot applies constrained optimization to select a waypoint that maximizes this estimated reward.

\p{Testing in Simulated and Real Environments}
We test our proposed approach across six simulated benchmark tasks, and two real-world robotic manipulation tasks.
The results suggest that our method leads to higher rewards and faster convergence than SAC or PPO baselines.

\section{Related Work} \label{sec:related}

We will introduce a reinforcement learning approach for robot manipulators \cite{han2023survey} that is based on trajectory waypoints.

\p{Hierarchical Reinforcement Learning}
Similar to our approach, work on hierarchical reinforcement learning breaks down a robot's behavior into high-level and low-level policies \cite{sutton1999between}.
The low-level policies are temporally extended actions (e.g., subtasks, options, goals), and the high-level policy connects these subtasks to complete the overall task.
For example, a robot arm can learn to insert a peg into a block by combining grasping and reaching subtasks \cite{nasiriany2022augmenting}.
The low-level policies can be parameterized (e.g., reaching for a given $xyz$ position) \cite{nachum2018data, nasiriany2022augmenting} or learned from scratch (e.g., policies conditioned on latent variables) \cite{bacon2017option, zhang2021hierarchical}.
In either case, hierarchical reinforcement learning has the potential to accelerate exploration by reducing the number of actions the high-level policy must take to explore its workspace.
Our approach can be viewed as an \textit{instance} of hierarchical reinforcement learning with high-level waypoints and a low-level controller that guides the robot to each waypoint.

\p{Motion Planning and Reinforcement Learning}
Within our proposed method we learn a trajectory to maximize reward in model-free settings. 
Along these lines, recent research has used motion planning as a building block towards larger manipulation tasks \cite{garrett2021integrated}.
In particular, we highlight existing methods that combine motion planning and reinforcement learning.
In some of these works reinforcement learning is used to move to nearby goals, and a motion planner connects these low-level waypoints into a complete task plan \cite{eysenbach2019search, gieselmann2021planning}.
Alternatively, in other works motion planning finds \textit{how} to reach a desired goal, and reinforcement learning determines \textit{where} to place these intermediate goals for the larger task \cite{xia2021relmogen, yamada2021motion}.
Prior research suggests that this combination of motion planning and reinforcement learning is effective for long-horizon tasks where the robot arm must precisely move to specific states (e.g., to pick up a block, the robot must first place its gripper directly above that block).

\p{Posterior Sampling for Reinforcement Learning}
Unlike these previous approaches, we will formulate the problem of learning the robot's trajectory as a sequence of multi-armed bandits, where each bandit corresponds to a waypoint.
One effective heuristic for solving multi-armed bandits is posterior sampling (i.e., Thompson sampling) \cite{russo2018tutorial}.
Posterior sampling has previously been extended to reinforcement learning,
with provable regret bounds in discrete state-action spaces \cite{osband2017posterior}.
However, this same approach is not tractable in continuous, high-dimensional environments.
Instead, recent research has developed accurate approximations using neural networks that maintain an \textit{ensemble} of learned models \cite{qin2022analysis, lee2021sunrise}.
We will similarly leverage an ensemble to approximate posterior sampling within our proposed algorithm.
\section{Problem Formulation} \label{sec:problem}

We consider settings in which a robot arm is given the reward function for a manipulation task, and the robot must learn to optimize this reward without relying on a model of the environment.
Model-free reinforcement learning approaches can already be applied to these problems settings; however, current approaches often take thousands to millions of roll-outs to learn benchmark tasks \cite{han2023survey, nasiriany2022augmenting}.
To learn these same tasks in fewer interactions, we will present a reinforcement learning approach where the robot arm moves between sequentially placed waypoints.

\p{MDP}
Reinforcement learning seeks to maximize the expected cumulative reward across a finite-horizon Markov Decision Process (MDP) $\mathcal{M} = \langle \mathcal{S}, \mathcal{A}, f, r, H, s^0 \rangle$. 
Here $s \in \mathcal{S}$ is the world state and $a \in \mathcal{A}$ is the robot arm's action.
Returning to our motivating example of opening a drawer, $s$ includes the robot's pose and the position and displacement of the drawer, and $a$ is the robot's end-effector velocity.
At each timestep $t$ the robot receives reward $r(s^t)$ and transitions between states based on the dynamics $s^{t+1} = f(s^t, a^t)$.
These dynamics are \textit{unknown}: the robot does not have access to a model of how its end-effector velocity will affect the displacement of the drawer or other environment variables.
By contrast, we assume that the reward function $r : \mathcal{S} \rightarrow \mathbb{R}$ is \textit{known}.
For instance, the robot observes $+1$ reward at states where the drawer is open (i.e., the drawer's displacement exceeds a threshold).

The robot repeatedly attempts to complete the same task in its environment and optimize the MDP $\mathcal{M}$. 
Each interaction (i.e., each episode) lasts for a total of $H$ timesteps.
Between interactions the world resets to the start state $s^0 \in \mathcal{S}$ (i.e., the drawer is closed and the robot returns to its home position). 
Let $K$ be the total number of interactions, i.e., the total number of times the robot can attempt to perform the task.

\p{Regret} 
During each interaction the system visits a trajectory of world states $\tau = \{s^0, \ldots, s^H\}$, such that $\tau^k$ is the system trajectory for the $k$-th interaction ($k \leq K$).
Remembering that $r(s)$ is the reward at a given state, we define $R(\tau) = \sum_{s \in \tau} r(s)$ as the total reward across trajectory $\tau$. 
Our objective is to identify trajectories that minimize regret. 
Consistent with prior works \cite{osband2016lower}, we define the regret as: 
\begin{equation} \label{eq:P1}
    REG(K) = \sum_{k = 1}^K R(\tau^*) - R(\tau^k)
\end{equation}
where $\tau^*$ is the best-case trajectory that maximizes the total reward (e.g., a trajectory that fully opens the drawer).
Intuitively, minimizing regret means the robot learns actions to complete the desired task in as few interactions as possible.
For model-free reinforcement learning to be practical on real-world robot arms, we need methods that quickly reduce the regret from \eq{P1} within a few interactions $K$.
\section{Reinforcement Learning with \\ Sequential Waypoints} \label{sec:method}

Our insight is that many manipulation tasks for robot arms can be broken down into a sequence of high-level waypoints that the arm must visit \cite{belkhale2023hydra, akgun2012keyframe, shi2023waypoint}.
Consider our running example of opening the drawer: here the robot arm (1) moves its end-effector to the handle, (2) grasps the handle, and then (3) pulls the drawer open.
Put another way, the robot arm completes the drawer task by following a trajectory between three high-level waypoints.
We will assume that the robot can move between the given waypoints using an existing controller for trajectory tracking.
Hence, our key challenge is determining \textit{where to place each subsequent waypoint} so that the robot's motion minimizes its regret.
In Section~\ref{sec:M1} we use our insight to reformulate reinforcement learning as a sequence of continuous multi-armed bandits. 
In Section~\ref{sec:MX} we list the assumptions behind this formulation, and then in Section~\ref{sec:M2} we explore the theoretical implications and derive lower bounds for regret. 
Finally, in Section~\ref{sec:M3} we present our algorithm for learning where to place each subsequent waypoint along the robot's trajectory.

\subsection{Reformulation as a Sequence of Multi-Armed Bandits} \label{sec:M1}

Introducing high-level waypoints lets us present a different formulation of the standard reinforcement learning problem described in Section~\ref{sec:problem}.
Under this new formulation the robot's high-level action is the choice of where to place the next waypoint (i.e., the next state of the robot arm).
As we will show, this leads to a sequence of continuous multi-armed bandits, where each bandit seeks to optimize the next waypoint along the robot's trajectory. 

\p{From Actions to Trajectories}
Let $s_\mathcal{R} \in \mathcal{S}_\mathcal{R}$ be the state of the robot arm.
The robot's state $s_\mathcal{R}$ is a subset of the world state $s$ such that $\mathcal{S}_\mathcal{R} \subseteq \mathcal{S}$.
For example, $s_\mathcal{R}$ could be the pose of the robot's end-effector and gripper, while $s$ could include $s_\mathcal{R}$ plus the position and displacement of the drawer.
The robot arm has direct control over state $s_\mathcal{R}$.
Put another way, the robot's end-effector velocity $a$ directly adjusts the position and orientation of its end-effector and gripper.

Under our proposed approach $s_\mathcal{R}$ is a \textit{waypoint}, and the robot arm completes its task by moving through a \textit{trajectory of waypoints}.
Consider our running example of opening a drawer: the robot could perform this task by moving to a state $s_\mathcal{R}^1$ where the robot grips the drawer handle, and then a state $s_\mathcal{R}^2$ where the robot arm moves away to pull the drawer open.
We therefore define the robot's reference trajectory $\xi$ as a sequence of desired waypoints: $\xi = \{s_\mathcal{R}^0, \ldots, s_\mathcal{R}^T\}$. 
The robot arm has a maximum of $T$ waypoints in its trajectory.
The choice of $T$ is left to the designer; however, we constrain $T < H$ so that the number of waypoints is less than the total number of timesteps in one interaction.
In what follows $\xi_i$ denotes a snippet of the trajectory $\xi$ with $i$ waypoints ($i \leq T$).
For example, $\xi_1$ is a trajectory snippet with just one waypoint $s_\mathcal{R}^1$, and $\xi_2$ is a trajectory snippet with two waypoints such that $\xi_2 = \{ s_\mathcal{R}^1, s_\mathcal{R}^2\}$.

Overall, $\xi$ defines a reference trajectory that we want the robot arm to follow. 
To track this trajectory we assume access to a low-level robot controller that takes actions $a$ to move the robot arm along $\xi$.
There are a variety of robot controllers for following reference trajectories \cite{spong2020robot}, and our approach is not tied to any specific controller choice.
In our experiments we use impedance control \cite{haddadin2016physical}.
In practice, impedance control causes the robot arm to linearly interpolate between the waypoints $s_\mathcal{R}$ while remaining compliant if the robot comes into contact with objects in the environment.

\p{From Unknown Dynamics to Unknown Rewards}
Given that the robot arm has chosen to roll-out trajectory $\xi$, we next want to determine how effectively this trajectory will perform the desired task.
As the robot tracks $\xi$ its low-level controller outputs actions $a$ (e.g., end-effector velocities, opening and closing the robot gripper).
These actions change not only the robot's state $s_\mathcal{R}$ but also the world state $s$.
Returning to our example, by moving the robot’s end-effector away from the drawer while holding the handle, the robot modifies the displacement of the drawer.
Accordingly, the complete trajectory of world states $\tau = \{s^0, \ldots, s^H\}$ is a function of the robot's trajectory $\xi$:
\begin{equation} \label{eq:M1}
    \tau = g(s^0, \xi)
\end{equation}
Here $g$ depends on both the unknown dynamics $f(s, a)$ and the robot's low-level controller.
The world starts in state $s^0$ and the robot takes low-level actions $a$ to follow trajectory $\xi$.
We overload this $g$ notation for trajectory snippets: $g(s^0, \xi_i)$ outputs the sequence of world states $s$ when the robot executes the trajectory snippet $\xi_i$.

The robot does not know how its own trajectory $\xi$ will map to changes in the world state $\tau$.
From our example: if the robot plans to reach some waypoint $s_\mathcal{R}^1$ and then another waypoint $s_\mathcal{R}^2$, the robot does not know beforehand whether this motion will open the drawer, push against the drawer, or miss the drawer altogether.
Without knowing the mapping $g$ the robot cannot anticipate what effects its trajectory will have on the world state; however, the robot can measure the reward for trajectories it has previously executed.
After the robot follows $\xi$ and observes the sequence of resulting world states $\tau$, the robot's total reward is: $R\big(g(s^0, \xi)\big) = R(\tau)$.
Similarly, after the robot executes a trajectory snippet $\xi_i$, its measured reward is $ R\big(g(s^0, \xi_i)\big)$.
We will write these trajectory rewards more compactly as:
\begin{equation} \label{eq:MX}
	R_\theta (s^0, \xi) = R\big(g(s^0, \xi)\big)
\end{equation}
where $\theta = \{\theta_1, \theta_2, \ldots, \theta_T\}$ is a set of \textit{reward parameters} induced by the unknown mapping $g$.
Here $\theta_1$ parameterizes to the reward for moving from the start state to waypoint $s_\mathcal{R}^1$, and $\theta_i$ parameterizes the reward for moving from waypoint $i-1$ to waypoint $i$.
The robot \textit{does not initially know these reward parameters}: returning to our example, at the first interaction the robot does not know how moving from $s_\mathcal{R}^{i-1}$ to $s_\mathcal{R}^i$ will affect the drawer's displacement.

\p{Multi-Armed Bandits}
Using waypoints $s_\mathcal{R}$ and reward parameters $\theta$, we can express our problem setting as a \textit{sequence of continuous multi-armed bandits} (MAB). 
Under this formulation the robot learns how to construct $\xi$ one waypoint at a time.
Each MAB in the sequence corresponds to a different waypoint: in the first MAB the robot learns waypoint $s_\mathcal{R}^1$, and in $i$-th MAB the robot learns waypoint $s_\mathcal{R}^i$.
More formally, each individual MAB $\mathcal{B}_i$ is defined by:
\begin{itemize}
    \item The continuous space of robot states $s_\mathcal{R} \in \mathcal{S}_\mathcal{R}$
    \item The continuous space of reward parameters $\theta_i \in \Theta_i$ for the transition from waypoint $i-1$ to waypoint $i$
    \item A prior over the reward parameters such that $\theta_i \sim P_i(\cdot)$
\end{itemize}
The robot operates within every MAB for a fixed number of interactions and then transitions to the next MAB. Below we overview both stages of this process for learning the robot's trajectory. We present a more detailed algorithm for solving these sequential MABs in Section~\ref{sec:M3}.

\p{Within MABs}
Imagine the robot is operating within MAB $\mathcal{B}_i$ (i.e., the robot is trying to determine where to place the $i$-th waypoint).
During an interaction the robot selects and plays a bandit arm.
More specifically, the robot selects the $i$-th waypoint $s_\mathcal{R}^i$ and executes the trajectory snippet $\xi_i$.
After the trajectory is complete, the robot measures the resulting reward $R_{\theta_i}(s^0, \xi_i)$ parameterized by the unknown weights $\theta_i$.
This cycle repeats at each interaction --- the robot tests a new choice of waypoint $s_\mathcal{R}^i$ and then observes the reward for the corresponding trajectory snippet $\xi_i$.

\p{Between MABs}
As the robot operates within MAB $\mathcal{B}_i$ it learns what waypoint $s_\mathcal{R}^i$ maximizes its measured reward.
We fix this waypoint in place when the robot transitions to the next MAB $\mathcal{B}_{i+1}$.
More generally, we \textit{freeze the strategy} the robot uses to choose $s_\mathcal{R}^i$ so that within MAB $\mathcal{B}_{i+1}$ the robot only explores waypoint $s_\mathcal{R}^{i+1}$.
To build trajectories with multiple waypoints the robot iterates through the previously learned strategies.
Returning to our example: perhaps in $\mathcal{B}_1$ the robot learned to reach the drawer and in $\mathcal{B}_2$ the robot learned to grasp the handle.
When the robot explores waypoint $s_\mathcal{R}^3$ in MAB $\mathcal{B}_3$, trajectory $\xi = \{s_\mathcal{R}^1, s_\mathcal{R}^2, s_\mathcal{R}^3\}$ will first move to the drawer ($s_\mathcal{R}^1$) and grasp the handle ($s_\mathcal{R}^2$).

\subsection{Assumptions} \label{sec:MX}

Before we analyze this proposed MAB formulation we want to clarify the assumptions behind our approach.
First, we assume that the robot arm can complete the desired task using a sequence of $T$ waypoints.
This assumption is violated when the designer chooses too few waypoints (i.e., the robot cannot open the drawer when $T=1$), or when the robot is faced with a task that does not break down into waypoints (i.e., the robot arm balancing an inverted pendulum).

Second, we assume that the reward $r(s)$ is composed of multiple local maxima, and each maxima corresponds to a stage of the task that can be completed by a single waypoint.
For example, the reward function for opening a drawer could includes terms for distance from the handle, whether the handle is grasped, and the displacement of the drawer.
This assumption is violated when the reward function does not encode a necessary subtask.
For example, if the robot needs to approach the drawer handle from above to open it --- but the reward only scores the distance to the center of the handle --- the robot will incorrectly place a waypoint at the center of the handle (and not approach it from above).

These assumptions limit the types of tasks for which our sequential MAB formulation applies, and also place an increased emphasis on reward design.
However --- as we will show in our experiments --- a variety of manipulation scenarios still satisfy the listed requirements.
We also note that automated reward design is an ongoing research topic, and in future work methods such as \cite{memarian2021self} could be leveraged to mitigate our second assumption.

\subsection{Lower Bounds on Regret} \label{sec:M2}

In Section~\ref{sec:M1} we reformulated reinforcement learning with waypoints as a sequence of MABs.
Here we explore the theoretical outcomes of this formulation.
Specifically, we compare lower bounds on regret --- as defined in \eq{P1} --- when using the standard MDP formulation and our special case MAB formulation.
These lower bounds correspond to ideal performance: i.e., if we identify optimal algorithms for both of the problem formulations, how quickly and efficiently will the robot learn to complete the task? 
To take advantage of existing analysis, we consider settings where both the state space $\mathcal{S}$ and the action space $\mathcal{A}$ are \textit{discrete}.
In our experiments (Sections~\ref{sec:sims} and \ref{sec:experiment}) we will test whether these theoretical results extend to continuous state-action spaces.

\p{Discrete Notation} 
Let $|\mathcal{A}|$ be the number of actions, let $|\mathcal{S}|$ be the number of world states, and let $|\mathcal{S}_\mathcal{R}|$ be the number of robot states. 
As a reminder, $H$ is the time horizon of one interaction and $K$ is the total number of interactions.
Within each MAB $\mathcal{B}_i$ the robot can select any state from $\mathcal{S}_\mathcal{R}$.
Hence, for one multi-armed bandit there are $|\mathcal{S}_\mathcal{R}|$ discrete arms.
Because we fix the chosen arm in MAB $\mathcal{B}_i$ when transitioning to MAB $\mathcal{B}_{i+1}$, across the sequence of $T$ MABs the robot reasons over a total of $T \cdot |\mathcal{S}_\mathcal{R}|$ discrete arms.

\p{Regret Bounds} 
Bubeck and Cesa-Bianchi \cite{bubeck2012regret} identified a lower bound on regret for any learning algorithm in an MAB.
When applied to our sequential MAB from Section~\ref{sec:M1}, this lower bound becomes: $REG(K) \geq \Omega\big(\sqrt{KT \cdot |\mathcal{S}_\mathcal{R}|}\big)$.

Other research has identified lower bounds on the regret for any reinforcement learning algorithm that repeatedly interacts with a finite-horizon MDP \cite{domingues2021episodic}.
When applied to our MDP $\mathcal{M}$ from Section~\ref{sec:problem}, this lower bound becomes: $REG(K) \geq \Omega\big(\sqrt{K H^2|\mathcal{S}||\mathcal{A}|}\big)$. Osband and Van Roy conjecture that this lower bound is unimprovable \cite{osband2016lower}.

These results quantify the lowest regret (i.e., the best performance) we could achieve under both problem formulations.
Put another way, these bounds tell us about the minimum number of interactions the robot will need before it can complete the underlying task.
It is not yet clear what algorithms will consistency reach these lower bounds.
However, comparing the lower bounds does provide theoretical justification for when we should formulate the problem as an MDP $\mathcal{M}$ (and solve using standard reinforcement learning methods) or as a sequence of MABs $\mathcal{B}$ (and solve using our proposed approach).
As $K \rightarrow \infty$, the lower bound for our MAB formulation is less than the MDP lower bound when:
\begin{equation} \label{eq:M3}
    T |\mathcal{S}_\mathcal{R}| < H^2 |\mathcal{S}||\mathcal{A}|
\end{equation}
We previously defined $T < H$ and $|\mathcal{S}_\mathcal{R}| \leq |\mathcal{S}|$.
Hence, \eq{M3} suggests that --- in scenarios where our assumptions from Section~\ref{sec:MX} apply --- the ideal performance when learning one waypoint at a time is better than the ideal performance for learning a robot policy.

\subsection{Approximate Solution with Posterior Sampling} \label{sec:M3}

In settings where the robot's task can be broken down into a sequence of waypoints, our theoretical analysis suggests that the MAB formulation from Section~\ref{sec:M1} is beneficial.
Here we propose an approximate solution to this sequential MAB that constructs the robot's trajectory one waypoint at a time.
Our approximation is based on \textit{posterior sampling}, a heuristic for multi-armed bandit problem \cite{russo2018tutorial}.
The core idea in posterior sampling is to maintain a distribution over the unknown reward parameters $\theta$; at each iteration, the robot samples a value of $\theta$ from the posterior and then rolls-out the bandit arm that maximizes reward $R_{\theta}$.
Based on the measured reward from this roll-out, the robot updates its distribution over $\theta$ and then repeats the process using the new posterior.
Unfortunately, we cannot directly apply posterior sampling because our state and parameter spaces are continuous and high-dimensional.
Instead, our algorithm approximates the posterior distribution through an \textit{ensemble} of reward models \cite{qin2022analysis, lee2021sunrise}.
Below we explain how we learn this ensemble within MAB $\mathcal{B}_i$, and also how we freeze the learned weights when transitioning to the next MAB $\mathcal{B}_{i+1}$.

\begin{algorithm}[t] 
\caption{Learning Waypoints via Posterior Sampling}
\label{alg:alg}
\begin{algorithmic}[1] 
\State Initialize buffer of reward models $\mathcal{R} = \{\}$\State Initialize waypoint index $i = 0$
\While {$i < T$}
    \State $i \gets i + 1$ \Comment{Transition to next waypoint}
    \State Initialize $N$ reward models with weights $\theta_{i,1}, \ldots \theta_{i, N}$
    \State Initialize dataset $\mathcal{D} \gets \{\}$
    \For {each interaction in MAB $\mathcal{B}_i$}
        \State Measure world start state $s^0$
        \State Initialize trajectory $\xi \gets \{\}$ \Comment{Build trajectory}
        \For {$(\theta_{j,1}, \ldots \theta_{j, N}) \in \mathcal{R}$}
            \State $s_\mathcal{R}^j \gets \text{arg}\max_{\mathcal{S}_\mathcal{R}} \dfrac{1}{N}\sum_{n=1}^N R_{\theta_{j, n}}(s^0, (\xi, s_\mathcal{R}))$
            \State $\xi \gets (\xi, s_\mathcal{R}^j)$ \Comment{Add previous waypoints}
        \EndFor
        \State Sample one or more indices $n \sim \text{uniform}(1, N)$
        \State $s_\mathcal{R}^i \gets \text{arg}\max_{\mathcal{S}_\mathcal{R}} R_{\theta_{i, n}}(s^0, (\xi, s_\mathcal{R}))$
        \State $\xi \gets (\xi, s_\mathcal{R}^i)$ \Comment{Add $i$-th waypoint}
        \State Roll-out trajectory $\xi$ in MDP $\mathcal{M}$
        \State Measure reward $R$ for dataset $\mathcal{D} \gets (s^0, \xi, R)$    
        \For {each reward model $n$}
            \State $\mathcal{L} \gets \sum_{(s^0, \xi, R)\in \mathcal{D}} \| R_{\theta_{i, n}}(s^0, \xi) - R \| $
            \State Update weights $\theta_{i, n}$ to minimize loss $\mathcal{L}$
        \EndFor
    \EndFor
    \State $\mathcal{R} \gets (\theta_{i,1}, \ldots \theta_{i, N})$ \Comment{Save learned reward models}
\EndWhile
\end{algorithmic}
\end{algorithm}

\p{Algorithm}
See Algorithm~\ref{alg:alg} for our proposed approach.
Our method is composed of two main parts to mirror the structure of sequential multi-armed bandits (Section~\ref{sec:M1}).
\textit{Within} a given MAB the robot approximates posterior sampling through an ensemble of reward models, and selects the next waypoint to maximize the estimated reward.
\textit{Between} MABs the robot saves these trained reward models; during future MABs the robot can refer back to the saved models to determine the previous waypoints along its trajectory.

\p{Within MABs} At the start of MAB $\mathcal{B}_i$ the robot initializes an ensemble of $N$ reward models (lines $4 - 6$).
These reward models estimate the unknown reward function $R_{\theta_i}(s^0, \xi_i)$ from \eq{MX}.
This ensemble of reward models approximates the posterior distribution over $\theta_i$.
The robot samples from this ensemble to estimate the reward for the current interaction (line $14$), and applies constrained optimization to identify the next waypoint $s_\mathcal{R}^i$ that maximizes the sampled reward (lines $15 - 16$).
The robot then rolls-out the resulting trajectory $\xi$ in the environment using its low-level controller (lines $17 - 18$).
We record the initial world state, the robot's trajectory of waypoints, and the measured reward (i.e., whether the drawer was opened during the interaction).
In lines $19 - 21$ the robot updates its posterior distribution over $\theta_i$ by training each reward model to match the measured rewards --- over repeated interactions, the ensemble of reward models should converge towards the true reward function $R_{\theta_i}(s^0, \xi_i)$.
This results in waypoints $s_\mathcal{R}^i$ which maximize the trajectory reward and complete the next stage of the task (i.e., reach for the handle or pull the drawer open).
In practice, sampling from an ensemble of models causes the robot to trade-off between exploring new waypoints and exploiting the best performing waypoints.

\p{Between MABs}
The process we have described so far determines the $i$-th waypoint along the trajectory.
But what about the previous waypoints $j=1$ to $j = i-1$?
After completing MAB $\mathcal{B}_{i-1}$ the robot saves the ensemble of reward models it has learned (line $24$)
When the robot moves on to the next MAB $\mathcal{B}_i$, we leverage these saved models at each interaction to find the previous waypoint $s_\mathcal{R}^{i-1}$.
Looking specifically at lines $10-12$, the robot loads each ensemble of models from its buffer (from $j = 1$ to $j = i-1$), and then applies constrained optimization to solve for waypoints 
$\xi = \{s_\mathcal{R}^1, \ldots s_\mathcal{R}^{i-1}\}$.
We emphasize that the reward models used to select these previous waypoints are frozen; the robot does not continue to train or modify the previous models.
Instead, the robot only focuses on the $i$-th waypoint while adding that new waypoint to a trajectory constructed across the sequence of previous MABs.

\p{Implementation Details}
Our experiments used an ensemble of $N=10$ reward models. Each model was a fully connected multi-layer perceptron with two hidden layers and a leaky ReLU activation function.
The reward models were updated by Adam with a learning rate of $0.001$ and MSE loss.
To find the waypoint that optimizes these reward models we applied constrained optimization --- in our experiments, we leveraged Sequential Least Squares Programming (SLSQP).
In practice this optimizer can get stuck in local maxima; to ameliorate this issue we used multiple initial seeds.
A repository of our code is available here: 
\url{https://github.com/VT-Collab/rl-waypoints}

\section{Benchmark Simulations} \label{sec:sims}

\begin{figure*}[]
    \centering
    \includegraphics[width=2\columnwidth]{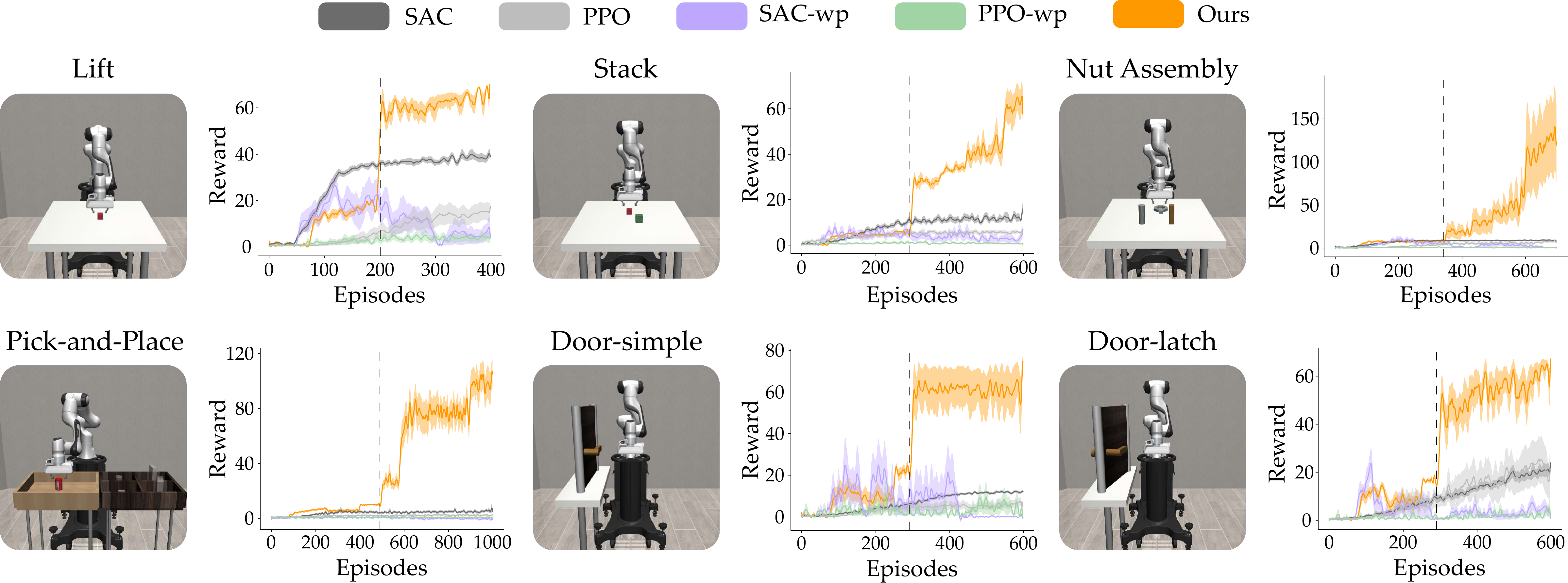}
    \caption{Simulation environments and rewards for six manipulation tasks. These benchmark tasks are taken from robosuite \cite{robosuite2020}. For each task the image on the left shows the environment setup, and the plot on the right shows the robot's rewards averaged over five runs. Higher rewards indicate better task performance. The dashed lines in the reward plots correspond to the episodes where \textbf{Our} approach added a new waypoint to the trajectory. Adding these new waypoints often causes a sharp increase in the robot's reward. This sudden change occurs because the new waypoint enables the robot to complete the next part of the task: e.g., in \textit{Lift} the robot needs one waypoint to grasp the block and a second waypoint to lift the block.}
    \label{fig:sim_lin}
    \vspace{-1 em}
\end{figure*}

In this section we evaluate how our proposed approach learns new manipulation tasks in simulated environments. 
We performed these benchmark experiments in \textit{robosuite}, a simulated robot environment with a set of standard manipulation tasks for robot arms \cite{robosuite2020}.
Across these benchmark tasks we compared our proposed approach to state-of-the-art reinforcement learning algorithms, and measured the rewards (i.e., the task performance) achieved by each method.

\p{Independent Variables}
We compared the performance of our proposed Algorithm~\ref{alg:alg} (\textbf{Ours}) to four baselines. First, we implemented soft actor-critic (\textbf{SAC}) \cite{haarnoja2018soft} and proximal policy optimization (\textbf{PPO}) \cite{schulman2017proximal}. Both \textbf{SAC} and \textbf{PPO} are model-free reinforcement learning algorithms that build a policy for selecting low-level robot actions $a$ based on state $s$. 

We also extended \textbf{SAC} and \textbf{PPO} to study their performance when using \textit{high-level waypoints} instead of \textit{low-level actions}. We will refer to these modified approaches as \textbf{SAC-wp} and \textbf{PPO-wp}. Similar to the original algorithms, both methods learn a policy that inputs world states $s$. But instead of outputting low-level actions $a$, now \textbf{SAC-wp} and \textbf{PPO-wp} output the next waypoint $s_\mathcal{R}^t$ that the robot should visit.
We applied the same impedance controller as in \textbf{Ours} to move the robot arm between waypoints.
The purpose of \textbf{SAC-wp} and \textbf{PPO-wp} was to test whether using high-level waypoints is the only advantage of our proposed approach: if \textbf{Ours} outperforms these baselines, this suggests that the framework we developed to learn the waypoints is also beneficial.

\p{Environment}
We used the benchmark tasks defined in robosuite \cite{robosuite2020} to evaluate the performance of each algorithm (see \fig{sim_lin}). In \textit{Lift} the robot needs to pick up a block. For the \textit{Stack} task the robot has to pick up one block and stack it on top of another block. In \textit{Nut Assembly} the robot picks up a nut and fits it on a peg. Next, in \textit{Pick-and-Place} the robot picks up different objects in the environment and places them in their respective containers. These four tasks all focused on the $xyz$ actions of the robot's end-effector. We also studied two manipulation tasks where the robot adjusts its position and orientation. In \textit{Door-simple} the robot needs to reach for the door latch and pull the door open. In the more complex \textit{Door-latch} task, the robot must turn the latch before it can open the door. We emphasize that the initial world state for each of these tasks was randomized at the start of every episode. For instance, the door was positioned at a different angle, and the blocks were placed at new locations.

\p{Procedure}
All the experiments were performed using a $7$-DoF Franka-Emika robot arm with impedance control in end-effector space. Each task had an episode length of $H = 100$ timesteps. Across all environments \textbf{Ours}, \textbf{SAC-wp} and \textbf{PPO-wp} were trained to complete the task using $T=2$ waypoints: the low-level controller used $50$ timesteps to reach waypoint $1$, and another $50$ timesteps to reach waypoint $2$. We made slight modifications to the given reward functions for the \textit{Stack}, \textit{Nut Assembly}, and \textit{Pick-and-Place} tasks by increasing the rewards for completing each stage of the task and adding penalties for knocking the objects over. 

\p{Dependent Variables}
The robot's task performance was measured using reward. If the robot encountered a trajectory of world state $\tau$ during a given episode, we reported: $R(\tau) = \sum_{s \in \tau} r(s)$. Higher rewards indicate better performance.

\p{Results}
We separated our results into two parts (see \fig{sim_lin} and Table~\ref{tab:rewards}). 
\fig{sim_lin} shows the rewards achieved by each model-free reinforcement learning algorithm as it trained in the environment.
Every interaction (i.e., every episode) lasted $H = 100$ timesteps, and we reported the total reward across that episode.
After the robot completed $K$ total interactions, we then saved the learned models and tested their performance.
Our results from these separate tests are listed in Table \ref{tab:rewards}.
Here the simulated robot attempted to complete each manipulation task $100$ times using the models it had learned from training.
Similar to training, during evaluation the position and orientation of objects in the environment was randomized at the start of every interaction.

Overall, none of the baselines were able to learn the tasks correctly within the limited number of episodes available for training. Across all tasks, we observed that \textbf{SAC} learned to reach and sometimes grasp the desired objects, while \textbf{PPO} only learned to reach for these objects. 
We also noticed that \textbf{SAC-wp} performed well at the start of training, but its performance dropped as the training progressed. We hypothesize that this may have occurred because --- instead of learning one waypoint at a time --- \textbf{SAC-wp} was simultaneously learning and adjusting both waypoints of the task. This may have prevented the robot from determining which waypoints were leading to higher rewards: e.g., did the robot succeed because of the position of waypoint 1 or waypoint 2?

Across both training and evaluation \textbf{Ours} outperformed the baselines. In \fig{sim_lin} we highlight that \textbf{Ours} had sudden jumps in episode reward. These jumps corresponded to episodes where the robot \textit{added a new waypoint} to its trajectory.
For example, in \textit{Lift} the robot arm solved the multi-armed bandit for waypoint $1$ during the first $200$ episodes, and then transitioned to the next MAB starting at episode $201$. The reward increased at this transition because the robot progressed to the next stage of the task.
Returning to \textit{Lift}, in the first $200$ episodes the robot learned one waypoint to reach and grasp the randomly initialized block. Starting from episode $201$, the robot applied what it had learned to grasp the block, and then explored its second waypoint to decide how to carry that block.
In summary, these simulation results suggest that \textbf{Ours} can efficiently learn a range of manipulation tasks in a limited number of interactions by breaking the task down into several waypoints, and learning the trajectory one waypoint at a time.

\begin{table}[]
\caption{
Evaluation results. After training in \fig{sim_lin}, we saved the learned models and evaluated their performance across $100$ episodes. We report average reward $\pm$ standard error. 
}
\label{tab:rewards}
\resizebox{\columnwidth}{!}{
\begin{tabular}{@{}llllll@{}}
\toprule
\textbf{Task}         & \multicolumn{4}{c}{\textbf{Mehtod}}                         & \textbf{}      \\ \midrule
\textbf{} &
  \multicolumn{1}{c}{\textbf{SAC}} &
  \multicolumn{1}{c}{\textbf{PPO}} &
  \multicolumn{1}{c}{\textbf{SAC-wp}} &
  \multicolumn{1}{c}{\textbf{PPO-wp}} &
  \multicolumn{1}{c}{\textbf{Ours}} \\
\textbf{Lift}         & $44.31 \pm 2.06$ & $17.58 \pm 0.34$ & $0.46 \pm 0.02$ & $0.22 \pm 0.06$ & $68.83 \pm 0.02$  \\
\textbf{Stack}        & $12.20 \pm 0.42$ & $3.04 \pm 0.19$  & $0.01 \pm 0.0$  & $0.86 \pm 0.09$ & $70.37 \pm 3.56$  \\
\textbf{Nut Assembly} & $9.51 \pm 0.14$  & $8.54 \pm 0.33$  & $0.10 \pm 0.01$ & $0.74 \pm 0.09$ & $190.14 \pm 11.7$ \\
\textbf{Pick-and-Place}   & $6.69 \pm 0.24$  & $1.37 \pm 0.10$  & $1.11 \pm 0.17$ & $0.24 \pm 0.05$ & $89.24 \pm 2.87$  \\
\textbf{Door-simple}  & $9.78 \pm 0.23$  & $4.58 \pm 0.20$  & $0.01 \pm 0$    & $1.67 \pm 2.59$ & $63.01 \pm 2.19$  \\
\textbf{Door-latch}   & $15.89 \pm 0.17$ & $6.64 \pm 0.26$  & $0.14 \pm 0.01$ & $0.29 \pm 0.03$ & $48.70 \pm 2.58$  \\ \bottomrule
\end{tabular}
}
\vspace{-1.em}
\end{table}
 
\section{Real-World Experiments} \label{sec:experiment}

\begin{figure}[t]
    \centering
    \includegraphics[width=\columnwidth]{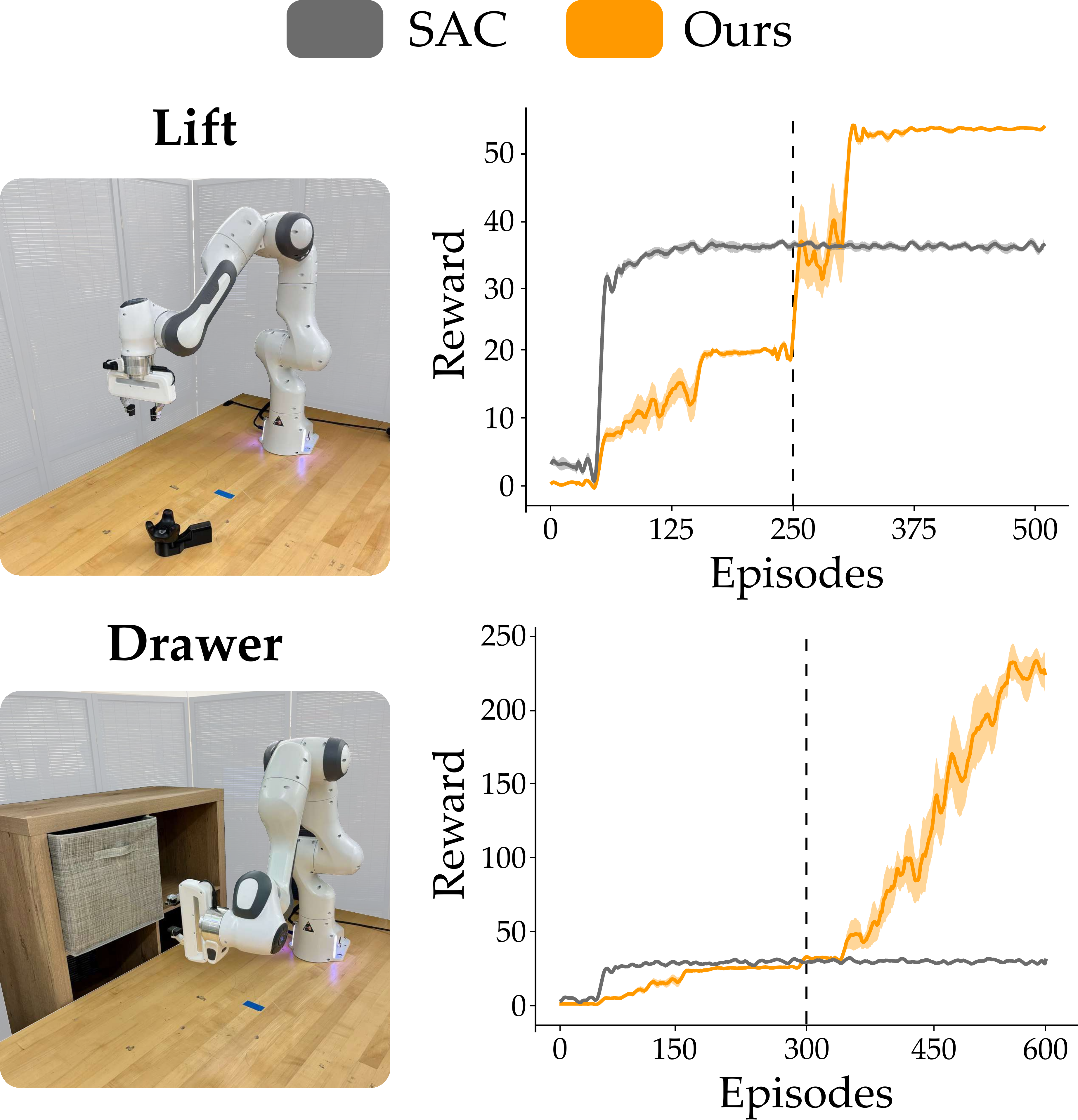}
    \caption{Setup and results from our real-world experiments in Section~\ref{sec:experiment}. In \textit{Lift} the robot learned to pick up an item, and in \textit{Drawer} it learned to open a drawer (also see \fig{front}). The robot measured the initial position of the item or drawer as a part of the start state $s^0$. \textbf{SAC} learned a policy that often moved to the object of interest, but did not correctly interact with that object. When using \textbf{Ours}, the robot built a trajectory of two waypoints: the first waypoint grasped the object, and then the second waypoint interacted with that object (e.g., picked the item up, pulled the drawer open). The dashed line in the reward plots corresponds to the episode where \textbf{Our} robot added a second waypoint to its trajectory.}
    \label{fig:exp}
    \vspace{-1 em}
\end{figure}

In this section we explore whether our proposed algorithm enables robot arms to learn manipulation tasks from scratch in real-world environments. We compare our approach (\textbf{Ours}) with the best-performing baseline from our simulations (\textbf{SAC}) across two different manipulation tasks. Within real-world environments with an actual robot arm, we find that Algorithm~\ref{alg:alg} leads to shorter training periods and more accurate task performance. 
See the robot's learned behaviors here: \url{https://youtu.be/MMEd-lYfq4Y}

\p{Experimental Setup} We conducted our real-world experiments on a $7$-DoF Franka Emika robot arm (see Figure~\ref{fig:exp}). We trained the robot to perform two different manipulation tasks: lifting an object (\textit{Lift}) and opening a drawer (\textit{Drawer}). The robot used \textbf{Ours} and \textbf{SAC} to attempt to learn each task. 

For both tasks the episodes lasted $H=140$ timesteps. The robot used end-effector velocity control to take actions within the environment. Under \textbf{SAC} the robot moved its end-effector and gripper based on the action output by the learned policy. With \textbf{Ours}, the robot learned two waypoints ($70$ timesteps each), and leveraged the same low-level controller to interpolate between its start state and the waypoints along its reference trajectory $\xi_\mathcal{R}$.

At the beginning of each episode the robot's initial position was constant, but the location of the object or drawer was uniformly randomized within a $6$-by-$6$ cm plane. 
We selected this smaller range to minimize the number of training episodes and prevent the robot from running into its joint limits.
We also applied boundary conditions to stop the robot from colliding with the table or attempting to move beyond its workspace. 
For \textit{Lift}, we used the same reward function as in robotsuite's lift task. 
For \textit{Drawer}, we defined a dense reward function that encouraged the robot to grasp the handle and then pull the drawer out in a straight line.

\p{Results} We first measured the episode rewards throughout training. Figure~\ref{fig:exp} shows the training rewards averaged across three runs for both tasks and methods. The maximum reward the robot achieved with \textbf{Ours} was about $1.3$ times \textbf{SAC} in \textit{Lift}, and about $7.75$ times \textbf{SAC} in \textit{Drawer}. 

After training completed, we used the robot's learned models to evaluate the robot's final success rate for both tasks. A robot trained with \textbf{Ours} grasped and lifted the randomly placed block in $16/20$ evaluation trials ($80\%$ success rate). Similarly, \textbf{Our} robot was able to grasp and at least partially open the drawer in $20/20$ evaluation trials. By contrast, under \textbf{SAC} the robot was never able to complete either task given the limited training budget.
\section{Conclusion} \label{sec:conclusion}

In this paper we focused on model-free reinforcement learning for robot arms.
We recognized that many of the everyday manipulation tasks that we want robots to learn can be broken down into a series of high-level waypoints (e.g., reaching an object).
We therefore proposed a framework for waypoint-based reinforcement learning, where the robot learns new tasks by building a trajectory one waypoint at a time.
Our key contribution was reformulating this problem as a sequence of multi-armed bandits: our theoretical analysis suggests that best-case solutions to this bandit formulation will outperform standard approaches.
We next introduced one possible algorithm for solving the sequential bandit problem.
Our proposed approach leveraged an ensemble of models to approximate posterior sampling: in practice, this method learned each subsequent waypoint to greedily maximize the robot's reward.
We found that our approach outperformed commonly used model-free reinforcement learning algorithms across a set of simulated and real-world manipulation tasks.
In future works we will explore more complicated tasks that require a higher number of waypoints.


\balance
\bibliographystyle{IEEEtran}
\bibliography{bibtex}

\begin{thebibliography}{10}
\providecommand{\url}[1]{#1}
\csname url@samestyle\endcsname
\providecommand{\newblock}{\relax}
\providecommand{\bibinfo}[2]{#2}
\providecommand{\BIBentrySTDinterwordspacing}{\spaceskip=0pt\relax}
\providecommand{\BIBentryALTinterwordstretchfactor}{4}
\providecommand{\BIBentryALTinterwordspacing}{\spaceskip=\fontdimen2\font plus
\BIBentryALTinterwordstretchfactor\fontdimen3\font minus
  \fontdimen4\font\relax}
\providecommand{\BIBforeignlanguage}[2]{{%
\expandafter\ifx\csname l@#1\endcsname\relax
\typeout{** WARNING: IEEEtran.bst: No hyphenation pattern has been}%
\typeout{** loaded for the language `#1'. Using the pattern for}%
\typeout{** the default language instead.}%
\else
\language=\csname l@#1\endcsname
\fi
#2}}
\providecommand{\BIBdecl}{\relax}
\BIBdecl

\bibitem{belkhale2023hydra}
S.~Belkhale, Y.~Cui, and D.~Sadigh, ``{HYDRA: H}ybrid robot actions for
  imitation learning,'' in \emph{Conference on Robot Learning}, 2023.

\bibitem{akgun2012keyframe}
B.~Akgun, M.~Cakmak, K.~Jiang, and A.~L. Thomaz, ``Keyframe-based learning from
  demonstration: Method and evaluation,'' \emph{International Journal of Social
  Robotics}, vol.~4, pp. 343--355, 2012.

\bibitem{shi2023waypoint}
L.~X. Shi, A.~Sharma, T.~Z. Zhao, and C.~Finn, ``Waypoint-based imitation
  learning for robotic manipulation,'' in \emph{Conference on Robot Learning},
  2023.

\bibitem{spong2020robot}
M.~W. Spong, S.~Hutchinson, and M.~Vidyasagar, \emph{Robot Modeling and
  Control}.\hskip 1em plus 0.5em minus 0.4em\relax John Wiley \& Sons, 2020.

\bibitem{haddadin2016physical}
S.~Haddadin and E.~Croft, ``Physical human--robot interaction,'' in
  \emph{Springer Handbook of Robotics}.\hskip 1em plus 0.5em minus 0.4em\relax
  Springer, 2016.

\bibitem{han2023survey}
D.~Han, B.~Mulyana, V.~Stankovic, and S.~Cheng, ``A survey on deep
  reinforcement learning algorithms for robotic manipulation,'' \emph{Sensors},
  vol.~23, no.~7, 2023.

\bibitem{sutton1999between}
R.~S. Sutton, D.~Precup, and S.~Singh, ``Between {MDPs and semi-MDPs:} {A}
  framework for temporal abstraction in reinforcement learning,''
  \emph{Artificial Intelligence}, pp. 181--211, 1999.

\bibitem{nasiriany2022augmenting}
S.~Nasiriany, H.~Liu, and Y.~Zhu, ``Augmenting reinforcement learning with
  behavior primitives for diverse manipulation tasks,'' in \emph{IEEE
  International Conference on Robotics and Automation}, 2022.

\bibitem{nachum2018data}
O.~Nachum, S.~S. Gu, H.~Lee, and S.~Levine, ``Data-efficient hierarchical
  reinforcement learning,'' in \emph{Advances in Neural Information Processing
  Systems}, 2018.

\bibitem{bacon2017option}
P.-L. Bacon, J.~Harb, and D.~Precup, ``The option-critic architecture,'' in
  \emph{AAAI}, 2017.

\bibitem{zhang2021hierarchical}
J.~Zhang, H.~Yu, and W.~Xu, ``Hierarchical reinforcement learning by
  discovering intrinsic options,'' in \emph{International Conference on
  Learning Representations}, 2021.

\bibitem{garrett2021integrated}
C.~R. Garrett, R.~Chitnis, R.~Holladay, B.~Kim, T.~Silver, L.~P. Kaelbling, and
  T.~Lozano-P{\'e}rez, ``Integrated task and motion planning,'' \emph{Annual
  Review of Control, Robotics, and Autonomous Systems}, 2021.

\bibitem{eysenbach2019search}
B.~Eysenbach, R.~R. Salakhutdinov, and S.~Levine, ``Search on the replay
  buffer: {B}ridging planning and reinforcement learning,'' in \emph{Advances
  in Neural Information Processing Systems}, 2019.

\bibitem{gieselmann2021planning}
R.~Gieselmann and F.~T. Pokorny, ``Planning-augmented hierarchical
  reinforcement learning,'' \emph{IEEE Robotics and Automation Letters},
  vol.~6, no.~3, pp. 5097--5104, 2021.

\bibitem{xia2021relmogen}
F.~Xia, C.~Li, R.~Mart{\'\i}n-Mart{\'\i}n, O.~Litany, A.~Toshev, and
  S.~Savarese, ``Relmogen: {I}ntegrating motion generation in reinforcement
  learning for mobile manipulation,'' in \emph{IEEE International Conference on
  Robotics and Automation}, 2021.

\bibitem{yamada2021motion}
J.~Yamada, Y.~Lee, G.~Salhotra, K.~Pertsch, M.~Pflueger, G.~Sukhatme, J.~Lim,
  and P.~Englert, ``Motion planner augmented reinforcement learning for robot
  manipulation in obstructed environments,'' in \emph{Conference on Robot
  Learning}, 2021.

\bibitem{russo2018tutorial}
D.~J. Russo, B.~Van~Roy, A.~Kazerouni, I.~Osband, and Z.~Wen, ``A tutorial on
  thompson sampling,'' \emph{Foundations and Trends in Machine Learning},
  vol.~11, no.~1, pp. 1--96, 2018.

\bibitem{osband2017posterior}
I.~Osband and B.~Van~Roy, ``Why is posterior sampling better than optimism for
  reinforcement learning?'' in \emph{International Conference on Machine
  Learning}, 2017.

\bibitem{qin2022analysis}
C.~Qin, Z.~Wen, X.~Lu, and B.~Van~Roy, ``An analysis of ensemble sampling,''
  \emph{Advances in Neural Information Processing Systems}, 2022.

\bibitem{lee2021sunrise}
K.~Lee, M.~Laskin, A.~Srinivas, and P.~Abbeel, ``Sunrise: {A} simple unified
  framework for ensemble learning in deep reinforcement learning,'' in
  \emph{International Conference on Machine Learning}, 2021.

\bibitem{osband2016lower}
I.~Osband and B.~Van~Roy, ``On lower bounds for regret in reinforcement
  learning,'' \emph{arXiv preprint arXiv:1608.02732}, 2016.

\bibitem{memarian2021self}
F.~Memarian, W.~Goo, R.~Lioutikov, S.~Niekum, and U.~Topcu, ``Self-supervised
  online reward shaping in sparse-reward environments,'' in \emph{IEEE/RSJ Int.
  Conf. on Intelligent Robots and Systems}, 2021.

\bibitem{bubeck2012regret}
S.~Bubeck and N.~Cesa-Bianchi, ``Regret analysis of stochastic and
  nonstochastic multi-armed bandit problems,'' \emph{Foundations and Trends in
  Machine Learning}, vol.~5, no.~1, pp. 1--122, 2012.

\bibitem{domingues2021episodic}
O.~D. Domingues, P.~M{\'e}nard, E.~Kaufmann, and M.~Valko, ``Episodic
  reinforcement learning in finite mdps: Minimax lower bounds revisited,'' in
  \emph{Algorithmic Learning Theory}, 2021.

\bibitem{robosuite2020}
Y.~Zhu, J.~Wong, A.~Mandlekar, R.~Mart\'{i}n-Mart\'{i}n, A.~Joshi,
  S.~Nasiriany, and Y.~Zhu, ``robosuite: A modular simulation framework and
  benchmark for robot learning,'' in \emph{arXiv:2009.12293}, 2020.

\bibitem{haarnoja2018soft}
T.~Haarnoja, A.~Zhou, P.~Abbeel, and S.~Levine, ``Soft actor-critic:
  {O}ff-policy maximum entropy deep reinforcement learning with a stochastic
  actor,'' in \emph{International Conference on Machine Learning}, 2018.

\bibitem{schulman2017proximal}
J.~Schulman, F.~Wolski, P.~Dhariwal, A.~Radford, and O.~Klimov, ``Proximal
  policy optimization algorithms,'' \emph{arXiv:1707.06347}, 2017.

\end{thebibliography}

\end{document}